# Aorta Segmentation for Stent Simulation

Jan Egger[1,2,3], Bernd Freisleben[1], Randolph Setser[4], Rahul Renapuraar[4], Christina Biermann[2,5], Thomas O'Donnell[3]

[1] Dept. of Math and Computer Science, U of Marburg, Hans-Meerwein-Str. 3, 35032 Marburg, Germany
[2] Siemens Healthcare, Computed Tomography, Siemensstr. 1, 91301 Forchheim, Germany
[3] Dept. of Imaging and Visualization, Siemens Corporate Research, 755 College Road East, Princeton, NJ 08540, USA
[4] Dept of Diagnostic Radiology, Cleveland Clinic Foundation, 9500 Euclid, Cleveland OH 44195
[5] Eberhard-Karls-Universität Tübingen, Radiologische Diagnostik, Hoppe-Seyler-Straße 3, 72076 Tübingen, Germany
{Jan.Egger.ext, tom.odonnell}@siemens.com

**Abstract.** Simulation of arterial stenting procedures prior to intervention allows for appropriate device selection as well as highlights potential complications. To this end, we present a framework for facilitating virtual aortic stenting from a contrast computer tomography (CT) scan. More specifically, we present a method for both lumen and outer wall segmentation that may be employed in determining both the appropriateness of intervention as well as the selection and localization of the device. The more challenging recovery of the outer wall is based on a novel minimal closure tracking algorithm. Our aortic segmentation method has been validated on over 3000 multiplanar reformatting (MPR) planes from 50 CT angiography data sets yielding a Dice Similarity Coefficient (DSC) of 90.67%.

**Keywords:** stent, simulation, vessel, segmentation, thrombus, tracking

## 1 Introduction

Aortic aneurysms, the irreversible widening of the aorta's outer wall, are the 13[th] leading cause of death in the US [1]. Within the last 10 years, treatment of this condition via stenting, known as EndoVascular Aortic Repair (EVAR), has become a clinical reality [2]. However, inappropriate stents may lead to endo-leakage (i.e., bleeding into the aneurismal sac due to in improper seal). EUROSTAR, the European consortium for EVAR evaluation, reported a 1% annual aortic rupture rate from EVAR procedures and an additional 2% conversion to open surgery due to stent failure [3]. Thus, selection of the proper stent is extremely important.

Simulation of stent placement enables clinicians to determine which devices are appropriate for an individual patient. In order to make this selection, recovery of the aorta is necessary.

Most aorta recovery methods focus on the lumen; few tackle the more challenging problem of recovering the outer wall. In CT angiography scans, this task is made difficult primarily due to the presence of thrombus, or clotted blood. Depending on its location, thrombus can be virtually indistinguishable from surrounding tissues, making the recovery of the outer wall akin to the so-called apparent contour problem (Fig. 1). To complicate matters, thrombus often appears in the presence of aneurysms where the vessel wall diameter may increase rapidly along the length of the vessel. These sudden changes make the detection of the outer wall much more complicated. Thus, naïve approaches to recovery are not effective.

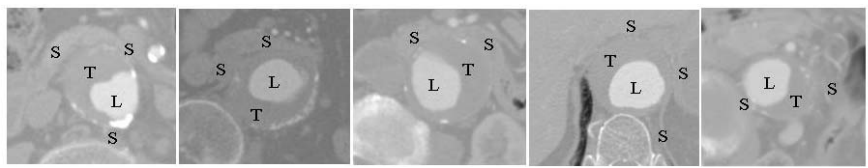

**Fig. 1.** Examples of CTA acquisitions showing lumen (L), thrombus (T) and surrounding tissues (S). Due to the similar intensity values of the thrombus and the surrounding tissue, the outer wall is often difficult to determine.

In this paper, we present a method for segmenting both the lumen and outer wall of the aorta that allows clinicians to (1) determine if intervention is warranted and if so, (2) which stenting device is most appropriate, and (3) where it should be positioned. By recovering the outer wall of the aorta, we are able to semi-automatically compute the maximal diameter of the vessel (the determining factor for intervention [1]) as well as isolate the extent of the lesion to insure total coverage). By recovering the lumen, we are able to make measurements (e.g., distance from lower renal artery to the iliac bifurcation, or the luminal diameter) in order to decide upon the appropriate device.

Briefly, we find the centerline of the aortic lumen and resample images orthogonal to its length. In each of the cross-sectional MPRs we isolate the lumen through a combination of thresholding, morphological operations and connected component analysis. Then, the user delineates the contours of the outer wall (semi-automatically) in three slices: proximal to the lesion (aneurysm), distal to it, and at some point in between. The outer wall in the remaining slices is segmented using a tracking scheme based on computing the minimal closure on a graph composed from two consecutive mprs one of which already contains a segmentation. The tracking takes place independently from each contour delineated by the user towards neighboring contours. A 3D active contour model of the stent is then employed to simulate the its placement.

The paper is organized as follows. Section 2 overviews related work. Section 3 presents our methods for aorta segmentation. In Section 4, experimental results are discussed. Section 5 concludes the paper and outlines areas for future work.

## 2   Related Work

In the area of stent planning and simulation for aortic aneurysms, we previously introduced a method for visualizing models of tube- and Y-stents placed virtually into preoperative CT data [4]. A drawback of this approach is that it only uses the lumen for the measurement and the simulation. Wong and Chung [5] present an approach for identifying different vascular abnormalities. However, real clinical data was only used for cerebral aneurysms. A method for simulating and visualizing non-bifurcated tube-stents in vascular CT data is presented by Florez-Valencia et al. [6]. However, Florez-Valencia et al.'s method also only uses the lumen for simulation.

Regarding the segmentation of the aorta's outer wall, De Bruijne et al. [7], apply an active shape model formulation in which landmarks may be defined by correlating with adjacent slices rather than training data. The model is initialized manually and their two-slice model climbs one slice at a time along the aorta. Their approach, however, requires a training set and runs the risk of degenerating modes of variation since the aortic cross-sections are frequently circular. By contrast, our tracking is based on an optimal s-t minimal cut.

Li et al. [8] present a global graph-based approach to segment the luminal and outer surfaces. In contrast to Li et al., our graph construction encodes the concept of directionality along the vessel path, and our node weights enforce both a fixed segmentation and a "forbidden" zone based on the previous MPR's segmentation. In addition, in Li et al.'s work, the entire length of vessel is recovered simultaneously as well as both the luminal and outer surfaces. We, on the other hand, recover the lumen in a separate operation, and *track* the adventitia from cross-sectional slice to cross-sectional slice. This allows us to introduce more sophisticated information (e.g., lumen morphology), and test multiple hypotheses as we proceed.

## 3   Methods

Our method begins by computing the luminal centerline of the vessel along a length of interest [9]. The resulting centerline determines a series of MPR planes that display the vessel cross-sections. The lumen on these MPRs is then automatically segmented using the approach of Ouvrard et al. [9]. Briefly, the image is smoothed with a Gaussian filter and is thresholded based on the sampling of the signal intensities along the luminal centerline. Morphological operations are then performed, and connected component analysis is applied. The component that intersects the lumen centerline is selected and refined by the removal of calcium and correction at branch points (detected by measuring ellipticity).

Then, the user either manually or through some automatic means, segments the outer wall on at least one MPR plane to initiate the tracking. (Additional contours may be provided by the user to support the overall automatic segmentation of the vessel.)

For outer wall recovery, it is possible, by tracking, to introduce more sophisticated information into the recovery process. For example, in some cases, lumen morphology can be used to predict outer wall location in the presence of severe thrombus. Specifically, elliptic lumen may provide clues to the radius of curvature of the outer wall cross-section (Fig. 2). Furthermore, by tracking we are able to test multiple hypotheses regarding the best fit outer wall before proceeding along the vessel.

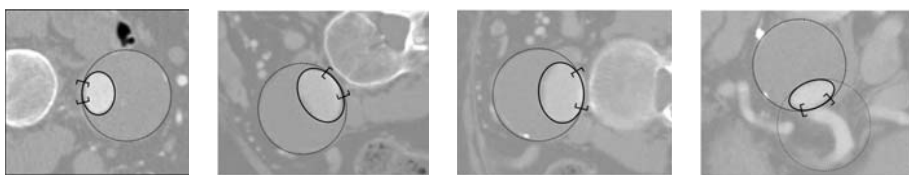

**Fig. 2.** The radius of curvature of the lumen in the bracketed region is very similar to that of the outer wall and can be exploited to recover the latter. Rightmost: Note that for each elliptic lumen there are two possible outer walls. We must determine which one is correct.

Tracking is performed between two consecutive MPRs: $M_0$ and $M_1$. Initially, $M_0$ contains a fixed segmented contour and the outer wall on $M_1$ is to be automatically segmented as explained below. Upon completion, the newly segmented MPR, $M_1$, is assigned to $M_0$ and the next consecutive MPR is assigned to $M_1$ and the process is repeated.

For each iteration of the tracking, $M_0$ and $M_1$ are used to construct a directed graph $G = (V, E)$ where $V$ is the set of nodes and $E$ is the set of edges. The nodes of $V$ are extracted from the MPRs by equidistantly sampling along rays that are projected from the point where the centerline intersects the MPR (Fig. 3). These nodes are arranged in a 3D lattice such that $x$ sweeps along the circumference of the vessel wall, $y$ from $M_0$ to $M_1$, and $z$ from the center of the vessel outwards (in other words, cylindrical coordinates where z is the radius, y the height, and x is theta). The assignment of nodal costs and edge architecture facilitate the tracking approach. More specifically, a closed set of a directed graph is defined as a subset of nodes such that all the successors of any node are also in the subset. The cost of a set is the sum of the costs of the nodes in the set. Our graph is constructed such that the computation of the closed set of minimal cost yields the optimal outer layer in $M_1$.

The cost $w(x, y, z)$ for every node $v \in V$ is assigned in the following manner: For $M_0$ we have a fixed segmentation $M_{fixed}(x, z)$. Therefore, $w(x, 0, z) = 0$ for $z \leq M_{fixed}(x, z)$ and $w(x, 0, z) = \infty$ for $z > M_{fixed}(x, z)$. On $M_1$, initially, weights are set to $c(x, 1, z)$ when z is zero and otherwise to $c(x, 1, z) - c(x, 1, z-1)$ where $c(x, y, z)$ is the absolute value of the intensity difference between an average thrombus grey value and the grey value of the voxel at position $(x, y, z)$. We then adjust these weights to enforce a "forbidden" zone on $M_1$ based on the fixed segmentation on $M_0$. That is, $w(x,1,z) = \infty$ if $z > M_{fixed}(x,z) + \Delta_p$ where $\Delta_p$ specifies the extent of the "forbidden" zone.

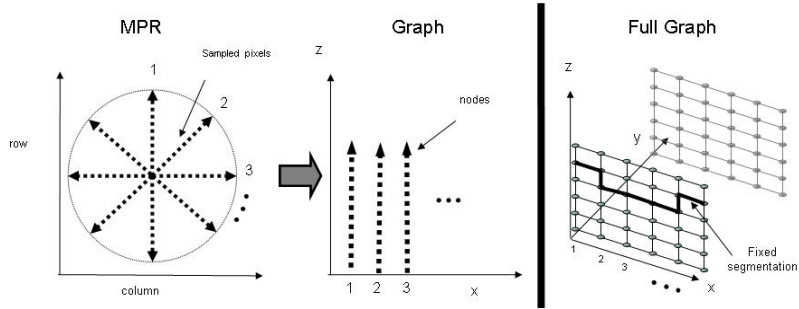

**Fig. 3.** Left: Pixels are sampled along rays sent out from the lumen center on a single MPR slice. Middle: Creation of an "unfolded" image. The rays become columns. Right: The full graph constructed from two unfolded slices. The fixed segmentation $M_{fixed}(x,z)$ is also shown.

The arcs $<v_i, v_j> \in E$ of the graph $G$ connect two nodes $v_i, v_j$. There are three types of arcs: $z$ arcs ($A_z$), $x$-$z$ arcs ($A_{xz}$), and $y$-$z$ arcs ($A_{yz}$) ($X$ is the number of rays sent out radially ($x = 0,...,X-1$), $Y$ the number of MPRs ($y = 0..1$) and $Z$, the number of sampled voxel along one ray $z = (0,...,Z-1)$):

$$A_z = \{\langle V(x,y,z), V(x,y,z-1)\rangle \mid z > 0\}$$
$$A_{xz} = \bigcup \begin{cases} \{\langle V(x,y,z), V(x+1,y,\max(0,z-\Delta_x))\rangle \mid x \in \{0,...,X-2\}\}, \\ \{\langle V(x,y,z), V(x-1,y,\max(0,z-\Delta_x))\rangle \mid x \in \{1,...,X-1\}\} \end{cases} \quad (1)$$
$$A_{yz} = \bigcup \begin{cases} \{\langle V(x,y,z), V(x,y+1,\max(0,z-\Delta_y))\rangle \mid y \in \{0,...,Y-2\}\}, \\ \{\langle V(x,y,z), V(x,y-1,\max(0,z-\Delta_y))\rangle \mid y \in \{1,...,Y-1\}\} \end{cases}$$

The intracolumn arcs $A_z$ ensure that all nodes below the surface in the graph are included to form a closed set (correspondingly, the interior of the vessel is separated from the exterior in the original MPR). The intercolumn arcs $A_{xz}$ and $A_{yz}$ constrain the set of possible segmentations and enforce smoothness via two parameters $\Delta_x$ and $\Delta_y$. The larger these parameters are, the greater the number of possible segmentations (see Figure 4).

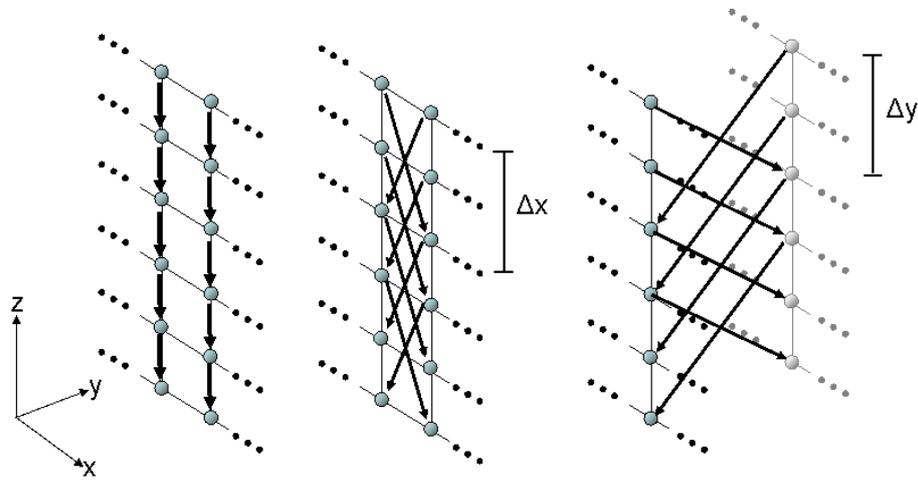

**Fig. 4.** Left: $A_z$ arcs. Middle: $A_{xz}$ arcs. Right: $A_{yz}$ arcs.

The minimal cost closed set on the graph is computed via a polynomial time s-t cut 10], creating an optimal segmentation of the outer wall of the vessel on $M_1$. This segmented contour serves as a starting point for the next iteration, where the distal slice becomes the proximal slice and the tracking continues. Thereby, we expect the segmentation to be correct for the proximal (manual) slice and permanently fix this slice by setting the costs for the unfolded positions in the graph that correspond to the (manually) segmented contour to zero. All other positions get a maximum cost value. Additionally, this proximal contour is used to set up a restricted search region (the "forbidden" zone) in the adjacent distal MPR plane. Only the voxels within this search region are used to create the directed graph for the distal MPR. In addition, the delta value that controls the stiffness between the slices $\Delta_y$ is set to a very low value ($<5$) to bind the slices very tightly to each other. This is possible because the outer contours between two adjacent MPRs are similar in their sizes, if the distance between these two slices along the centerline is small ($\leq 5mm$).

When the vessel lumen is found to be smoothly elliptical, it frequently follows that the outer wall shares the radius of curvature along the major axis. We exploit this in the tracking by predicting a more accurate center of the vessel. Then, that MPR can be unfolded again with a more accurate center leading to more accurate segmentations. For each elliptical lumen, there are two possible vessel centers $(c_x, c_z)$ corresponding to the two sides along the major axis (Fig. 2 Rightmost). To determine which is the correct center, we integrate over a small square of dimension $b$ centered on the two candidates:

$$\int_{-b/2}^{b/2} \int_{-b/2}^{b/2} T(c_x + x, c_z + z) dx dz \qquad (2)$$

where $T(x,z) = 0$ if $(x,z)$ is lumen, otherwise it is the voxel intensity value. The correct center should be sitting in a patch of thrombus and therefore have a higher value for this integral. Thus, we select the larger of the two values as indicating as the correct center.

Tracking can be initiated from different points along the vessel. For example, the user may manually segment at two points and begin the tracking from each point towards the middle. This can be extended to multiple initiation points where tracking follows in the direction towards the half-way point to the next manually segmented MPR (in both proximal and distal directions, if possible). More generally, for the case of 3 initiation points, given *n* total MPRs along the centerline $M_i$ ($i = 0,...,n-1$), with the first proximal MPR $M_0$, the last distal MPR $M_{n-1}$ and a MPR $M_j$, with $0 < j < n-1$ and $n > 2$. If $G_i$ is the 2D Graph that belongs to the MPR $M_i$, and $G_i G_{i+1}$ describes the construction of a thin 3D graph, tracking initiates from the graphs $G_0 G_1$, $G_{n-1} G_{n-2}$, $G_j G_{j-1}$, and $G_j G_{j+1}$, and all graphs that are used for the tracking can be described by the following equation:

$$G_k^{3D} = \begin{cases} G_l G_{l+1} & \text{if } (l \geq 0 \text{ and } l < \frac{j}{2}) \text{ or } (l \geq j \text{ and } l < \frac{n-j}{2} + j) \\ G_l G_{l-1} & \text{if } (l \leq j \text{ and } l \geq \frac{j}{2}) \text{ or } (l \leq n \text{ and } l \geq \frac{n-j}{2} + j) \end{cases} \quad (3)$$

## 4  Results

Stent simulation was performed on six pairs (pre-op/op) data sets as shown in Figure 5. Due to the scarcity of (pre-op/op) data sets, we additionally validated our segmentation on fifty clinical aorta data sets – containing overall more than 3000 MPR slices – with variations in anatomy and location of the pathology. Two trained observers performed manual segmentations of the data sets. The results were evaluated by calculating the Dice Similarity Coefficient (DSC) [11] (see Table 1). Figure 6 shows parts of the segmentation result of a pre-operative data set.

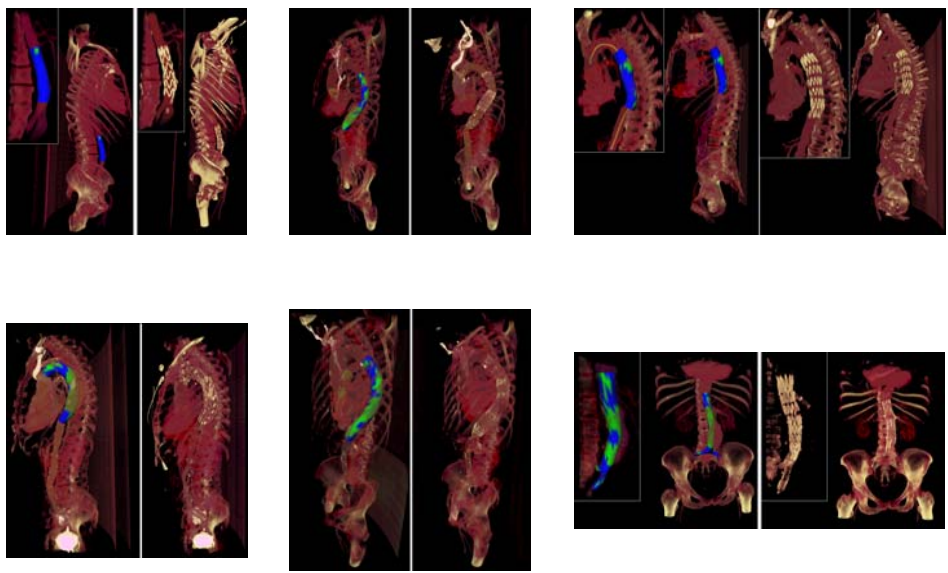

**Fig. 5.** Results of the stent simulation within six pre-operative data sets (left) compared with the corresponding post-operative data sets. The color coding indicates goodness of fit of simulated stent model to the wall. Green implies a lose fit, blue a tight fit.

Manual segmentation by a trained observer took about 20-40 minutes, automatic segmentation of a dataset with our method took about 30-90 seconds on an Intel Pentium 4 CPU, 2.8 GHz, 2 GB RAM. However, additional editing on some MPR slices was required, but these edits were achieved quite quickly because the automatic segmentation provides a border that at least fits partially to the desired contour.

**Table 1.** Summary of results: min., max., mean $\mu$ and standard deviation $\sigma$ for 50 scans.

|  | Area of clot (cm$^3$) | | Number of Voxels | | DSC (%) |
|---|---|---|---|---|---|
|  | manual | Automatic | manual | automatic |  |
| Min | 79.58 | 75.36 | 24440 | 23168 | 72.80 |
| Max | 820.21 | 808.41 | 3005020 | 3130590 | 99.27 |
| $\mu \pm \sigma$ | 342.95 ± 212.59 | 347.86 ± 205.44 | 452820.3 | 443818.5 | 90.67 ± 5.31 |

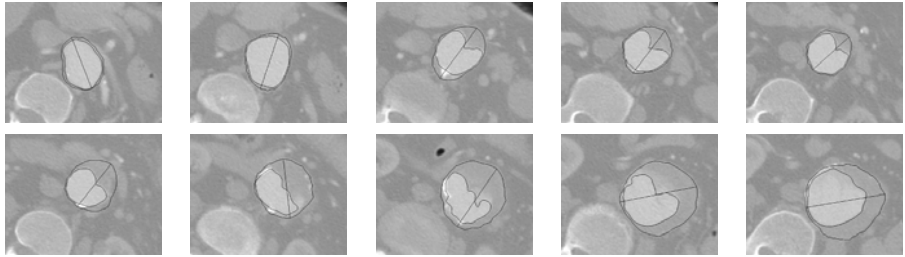

**Fig. 6.** Segmentation results for a pre-operative data set.

## 5 Conclusions

In this paper, we have presented a segmentation technique to facilitate aortic stent simulation. Our approach is effective in the presence of thrombus and is based on a novel iterative optimal graph-based algorithm that tracks along the centerline of the vessel. Using our approach, it is possible to perform an analysis to determine if intervention is required and acquire the measurements necessary to select the proper stenting device.

While our segmentation method is in general quite robust challenges remain. The diaphragm, for example, has signal intensity characteristics very similar to that of thrombus. In this region, the method may confuse the two. In addition, there are anatomical anomalies that can potentially cause problems. For example, in the presence of a dissection (a tear in the inner lining of the aorta causing two or more independent channels of blood) the segmentation of the outer wall may incorrectly exclude one of the channels (false lumen).

There are several areas of future work. For example, we want to segment the stents from the post-operative scans and calculate the DSC with the virtual stent models from the pre-operative scans.